\RequirePackage{snapshot}
\documentclass[10pt,twocolumn,letterpaper]{article}

\usepackage[utf8]{inputenc}
\usepackage[T1]{fontenc}
\usepackage{cvpr}
\usepackage{times}
\usepackage{epsfig}
\usepackage{graphicx}
\usepackage{amsmath}
\usepackage{amssymb}
\usepackage[caption=false]{subfig}
\usepackage{booktabs}

\usepackage[pagebackref=true,breaklinks=true,letterpaper=true,colorlinks,bookmarks=false]{hyperref}

\cvprfinalcopy 

\def\cvprPaperID{182} 

\ifcvprfinal\fi

\newcommand{\figref}[1]{Figure~\ref{#1}}
\newcommand{\tblref}[1]{Table~\ref{#1}}
\newcommand{\secref}[1]{Section~\ref{#1}}
\renewcommand{\eqref}[1]{Equation~(\ref{#1})}

\newcommand{\ckp}[2]{$CK_{#1}P_{#2}$}
\newcommand{\cext}{\ckp{8}{16}$ext$}
\newcommand{\cfin}{$FK_{8}P_{8}$}
\newcommand{\cray}{\ckp{8}{8}$ray$}
\newcommand{\csin}{$SK_{8}P_{8}$}

\begin{document}

\title{PupilNet: Convolutional Neural Networks for Robust Pupil Detection}



\author{
	Wolfgang Fuhl\\
	Perception Engineering Group \\
	University of T\"ubingen \\
	{\tt\small
	wolfgang.fuhl@uni-tuebingen.de
	}
	\and
	Thiago Santini\\
	Perception Engineering Group \\
	University of T\"ubingen \\
	{\tt\small
	thiago.santini@uni-tuebingen.de
	}
	\and
	Gjergji Kasneci\\
	SCHUFA Holding AG\\
	{\tt\small
	\phantom{en} 
	gjergji.kasneci@schufa.de
	\phantom{.de}
	}
	\and
	Enkelejda Kasneci\\
	Perception Engineering Group \\
	University of T\"ubingen \\
	{\tt\small
	enkelejda.kasneci@uni-tuebingen.de
	}
}


\maketitle

\begin{abstract}

Real-time, accurate, and robust pupil detection is an essential prerequisite for
pervasive video-based eye-tracking.
However, automated pupil detection in real-world scenarios has proven to be an
intricate challenge due to fast illumination changes, pupil occlusion, non
centered and off-axis eye recording, and physiological eye characteristics.
In this paper, we propose and evaluate a method based on a novel dual
convolutional neural network pipeline.
In its first stage the pipeline performs coarse pupil position identification using a
convolutional neural network and subregions from a downscaled input image to
decrease computational costs.
Using subregions derived from a small window around the initial pupil position
estimate, the second pipeline stage employs another
convolutional neural network to refine this position, resulting in an increased pupil detection rate up to 25\% in comparison with the best performing
state-of-the-art algorithm.
Annotated data sets can be made available upon request.


\end{abstract}

\section{Introduction}

For over a century now, the observation and measurement of eye movements have
been employed to gain a comprehensive understanding on how the human oculomotor
and visual perception systems work, providing key insights about cognitive
processes and behavior~\cite{wade2005moving}. Eye-tracking devices are rather
modern tools for the observation of eye movements.
In its early stages, eye tracking was restricted to static activities, such as
reading and image perception~\cite{yarbus1957perception}, due to restrictions
imposed by the eye-tracking system -- e.g., size, weight, cable connections, and
restrictions to the subject itself.
With recent developments in video-based eye-tracking technology, eye tracking
has become an important instrument for cognitive behavior studies in many areas,
ranging from real-time and complex applications (e.g., driving
assistance based on eye-tracking input~\cite{kasneci2013towards} and gaze-based
interaction~\cite{turner2013eye}) to less demanding use cases, such as usability
analysis for web pages~\cite{cowen2002eye}.
Moreover, the future seems to hold promises of pervasive and unobtrusive
video-based eye tracking~\cite{kassner2014pupil}, enabling research and
applications previously only imagined.
%

\begin{figure}[h]
	\begin{center}
		\subfloat[][\label{a}]{
			\includegraphics[width=.23\columnwidth]{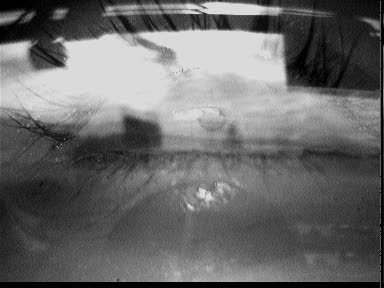}
			\label{fig:reflection}
		}
		\subfloat[][\label{b}]{
			\includegraphics[width=.23\columnwidth]{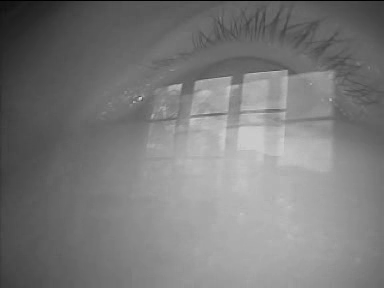}
			\label{fig:reflection2}
		}
		\subfloat[][\label{c}]{
			\includegraphics[width=.23\columnwidth]{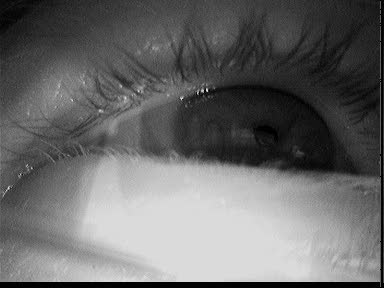}
			\label{fig:dark}
		}
		\subfloat[][\label{d}]{
			\includegraphics[width=.23\columnwidth]{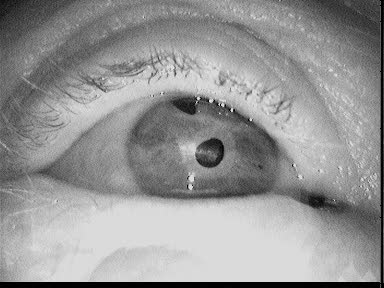}
			\label{fig:physiological}
		}
		\caption{
			Images of typical pupil detection challenges in
			real-world scenarios: (a) and (b) reflections, (c) pupil
			located in dark area, and (d) unexpected physiological structures.
		}
	\end{center}
\end{figure}

While video-based eye tracking has been shown to perform satisfactorily under
laboratory conditions, many studies report the occurrence of difficulties and
low pupil detection rates when these eye trackers are employed for tasks in
natural environments, for instance
driving~\cite{kasneci2013towards,liu2002real,trosterer2014eye} and
shopping~\cite{kasneci2014homonymous}.
The main source of noise in such realistic scenarios is an unreliable pupil
signal, mostly related to intricate challenges in the image-based pupil
detection.
A variety of difficulties occurring when using such eye trackers, such as
changing illumination, motion blur, and pupil occlusion due to eyelashes, are
summarized in~\cite{schnipke2000trials}.
Rapidly changing illumination conditions arise primarily in tasks where the
subject is moving fast (e.g., while driving) or rotates relative to unequally
distributed light sources, while motion blur can be caused by the image sensor
capturing images during fast eye movements such as saccades.
Furthermore, eyewear (e.g., spectacles and contact lenses) can result in
substantial and varied forms of reflections (\figref{fig:reflection} and
\figref{fig:reflection2}), non-centered or off-axis eye position relative to the
eye-tracker can lead to pupil detection problems, e.g., when the pupil is
surrounded by a dark region (\figref{fig:dark}).
Other difficulties are often posed by physiological eye characteristics, which
may interfere with detection algorithms (\figref{fig:physiological}).
As a consequence, the data collected in such studies must be post-processed
manually, which is a laborious and time-consuming procedure.
Additionally, this post-processing is impossible for real-time applications that
rely on the pupil monitoring (e.g., driving or surgery assistance).
Therefore, a real-time, accurate, and robust pupil detection is an essential
prerequisite for pervasive video-based eye-tracking.

State-of-the-art pupil detection methods range from relatively simple methods
such as combining thresholding and mass center
estimation~\cite{perez2003precise} to more elaborated methods that attempt to
identify the presence of reflections in the eye image and apply
pupil-detection methods specifically tailored to handle such challenges~\cite{fuhl2015excuse} -- a
comprehensive review is given in \secref{sec:related}.
Despite substantial improvements over earlier methods in real-world scenarios,
these current algorithms still present unsatisfactory detection rates in many
important realistic use cases (as low as 34\%~\cite{fuhl2015excuse}).
However, in this work we show that carefully designed and trained convolutional
neural networks~(CNN)~\cite{domingos2012few,lecun1998gradient}, which rely on
statistical learning rather than hand-crafted heuristics, are a substantial step
forward in the field of automated pupil detection.
CNNs have been shown to reach human-level performance on a multitude of pattern
recognition tasks (e.g., digit recognition~\cite{ciresan2012multi}, image
classification~\cite{krizhevsky2012imagenet}). 
These networks attempt to emulate the behavior of the visual processing system
and were designed based on insights from visual perception research.

We propose a dual convolutional neural network pipeline for image-based pupil detection. The first pipeline stage employs a shallow CNN on subregions of a downscaled version of the input image to quickly infer a coarse
estimate of the pupil location. This coarse estimation allows the
second stage to consider only a small region of the original image, thus, mitigating the impact of noise and decreasing computational costs.
The second pipeline stage then samples a small window around the coarse position
estimate and refines the initial estimate by evaluating
subregions derived from this window using a second CNN. We have focused on
robust learning strategies (batch learning) instead of more accurate ones
(stochastic gradient descent)~\cite{lecun2012efficient} due to the fact that an
adaptive approach has to handle noise (e.g., illumination, occlusion,
interference) effectively.

The motivation behind the proposed pipeline is (i) to reduce the noise in the coarse estimation of the pupil position, (ii) to reliably detect the exact pupil position from the initial estimate, and (iii) to provide an efficient method that can be run in real-time on hardware architectures without an accessible GPU.

In addition, we propose a method for generating training data in an online-fashion, thus being applicable to the task of pupil center detection in online scenarios. We evaluated the performance of different CNN configurations both in terms of quality and efficiency and report considerable improvements over stat-of-the-art techniques.

\section{Related Work}
\label{sec:related}

During the last two decades, several algorithms have addressed image-based pupil detection.
P\'{e}rez~et~al.~\cite{perez2003precise} first threshold the image and compute the mass center of the resulting dark pixels. This process is iteratively repeated in an area
around the previously estimated mass center to determine a new mass center until convergence.
The Starburst algorithm, proposed by Li et al.~\cite{li2005starburst}, first
removes the corneal reflection and then locates pupil edge points using an iterative feature-based approach.
Based on the RANSAC algorithm~\cite{fischler1981random}, a
best fitting ellipse is then determined, and the final ellipse parameters are determined by applying a model-based optimization.
Long~et~al.~\cite{long2007high} first downsample the image and search there for an approximate pupil location. The image area around this location is further processed and a parallelogram-based symmetric mass center algorithm is
applied to locate the pupil center.
In another approach, Lin~et~al.~\cite{lin2010robust} threshold the image, remove
artifacts by means of morphological operations, and apply inscribed
parallelograms to determine the pupil center.
Keil~et~al.~\cite{keil2010real} first locate corneal reflections; afterwards,
the input image is thresholded, the pupil blob is searched in the adjacency of
the corneal reflection, and the centroid of pixels belonging to the blob is
taken as pupil center.
Agustin~et~al.~\cite{san2010evaluation} threshold the input image and extract
points in the contour between pupil and iris, which are then fitted to an
ellipse based on the RANSAC method to eliminate possible outliers.
\'{S}wirski~et~al.~\cite{swirski2012robust} start with
a coarse positioning using Haar-like features. The intensity histogram of the coarse position is clustered using  k-means clustering, followed by a modified RANSAC-based ellipse fit.
The above approaches have shown good detection rates and robustness  in controlled settings, i.e., laboratory conditions.

Two recent methods, SET~\cite{javadi2015set} and ExCuSe~\cite{fuhl2015excuse},
explicitly address the aforementioned challenges associated with pupil detection
in natural environments. SET~\cite{javadi2015set} first extracts pupil pixels
based on a luminance threshold. The resulting image is then segmented, and the
segment borders are extracted using a Convex Hull method.
Ellipses are fit to the segments based on their sinusoidal components, and the
ellipse closest to a circle is selected as pupil.  ExCuSe~\cite{fuhl2015excuse}
first analyzes the input images with regard to reflections based on intensity
histograms. Several processing steps based on edge detectors, morphologic
operations, and the Angular Integral Projection Function are then applied to
extract the pupil contour. Finally, an ellipse is fit to this line using the
direct least squares method.

Although the latter two methods report substantial improvements over earlier
methods, noise still remains a major issue.
Thus, robust detection, which is critical in many online real-world
applications, remains an open and challenging problem~\cite{fuhl2015excuse}.

\section{Method}

\begin{figure}[h]
	\begin{center}
		\includegraphics[width=1.0\linewidth]{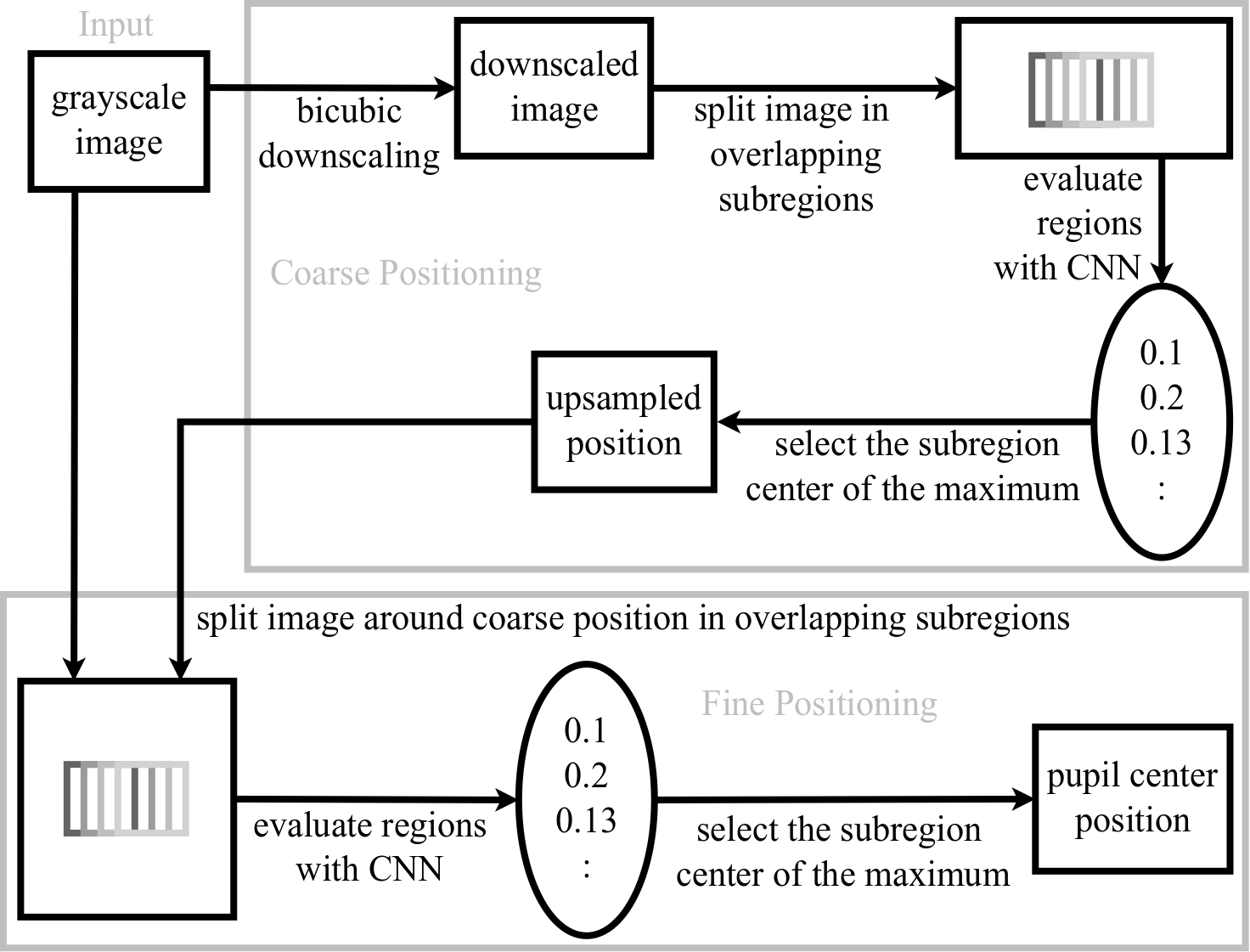}
	\end{center}
	\caption{
		Workflow of the proposed algorithm. First a CNN is employed to estimate
		a coarse pupil position based on subregions from a downscaled version of
		the input image. This position is then refined using subregions around
		the coarse estimation in the original input image by a second CNN.
	}
	\label{fig:workflow}
\end{figure}

The overall workflow for the proposed algorithm is shown in
\figref{fig:workflow}.
In the first stage, the image is downscaled and divided into overlapping
subregions. These subregions are evaluated by the first CNN, and the center of
the subregion that evokes the highest CNN response is used as a coarse pupil
position estimate.
Afterwards, this initial estimate is fed into the second pipeline stage. In this
stage, subregions surrounding the initial estimate of the pupil position in the
original input image are evaluated using a second CNN. The center of the
subregion that evokes the highest CNN response is chosen as the final pupil
center location.
This two-step approach has the advantage that the first step (i.e., coarse
positioning) has to handle less noise because of the bicubic downscaling of the
image and, consequently, involves less computational costs than detecting the
pupil on the complete upscaled image.

In the following subsections, we delineate these pipeline stages and their CNN
structures in detail, followed by the training procedure employed for each CNN.

\subsection{Coarse Positioning Stage}
\label{subsec:coarsestage}

\begin{figure}[h]
	\begin{center}
		\subfloat[][\label{a}]{
			\includegraphics[width=.3\columnwidth]{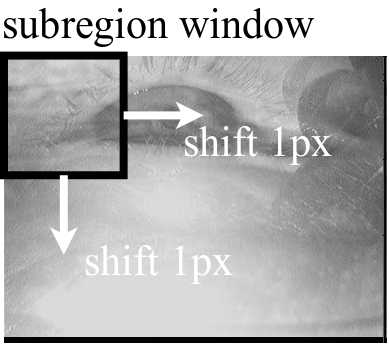}
			\label{fig:splitt}
		}
		\hfill
		\subfloat[][\label{b}]{
			\includegraphics[width=.57\columnwidth]{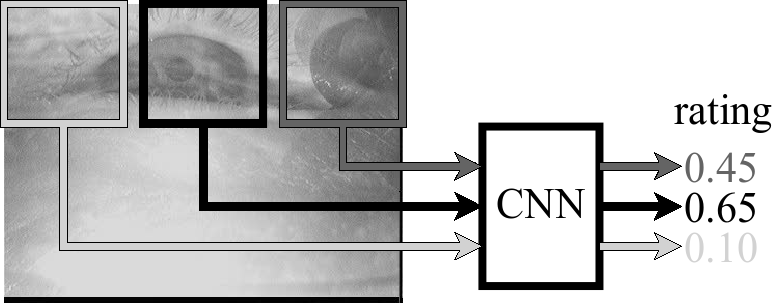}
			\label{fig:cnncoarse}
		}
		\caption{
			The downscaled image is divided in subregions of size $24\times24$
			pixels with a stride of one pixel (a), which are then rated by the
			first stage CNN (b).
		}
	\end{center}
\end{figure}

The grayscale input images generated by the mobile eye tracker used in this work
are sized $384\times288$ pixels.
Directly employing CNNs on images of this size would demand a large amount of
resources and would thus be computationally expensive, impeding their usage in
state-of-the-art mobile eye trackers. Thus, one of the purposes of the first
stage is to reduce computational costs by providing a coarse estimate that can
in turn be used to reduce the search space of the exact pupil location.
However, the main reason for this step is to reduce noise, which can be induced
by different camera distances, changing sensory systems between head-mounted eye
trackers~\cite{boie1992analysis,dussault2004noise,reibel2003ccd}, movement of the
camera itself, or the usage of uncalibrated cameras (e.g., focus or white
balance).
To achieve this goal, first the input image is downscaled using a bicubic
interpolation, which employs a third order polynomial in a two dimensional space
to evaluate the resulting values.
In our implementation, we employ a downscaling factor of four times, resulting
in images of $96\times72$ pixels.
Given that these images contain the entire eye, we chose a CNN input size of
$24\times24$ pixels to guarantee that the pupil is fully contained within a
subregion of the downscaled images.
Subregions of the downscaled image are extracted by shifting a $24\times24$
pixels window with a stride of one pixel (see \figref{fig:splitt}) and evaluated
by the CNN, resulting in a rating within the interval [0,1]
(see~\figref{fig:cnncoarse}).
These ratings represent the confidence of the CNN that the pupil center is
within the subregion. Thus, the center of the highest rated subregion is chosen
as the coarse pupil location estimation.

\begin{figure}[h]
	\begin{center}
		\includegraphics[width=1.0\linewidth]{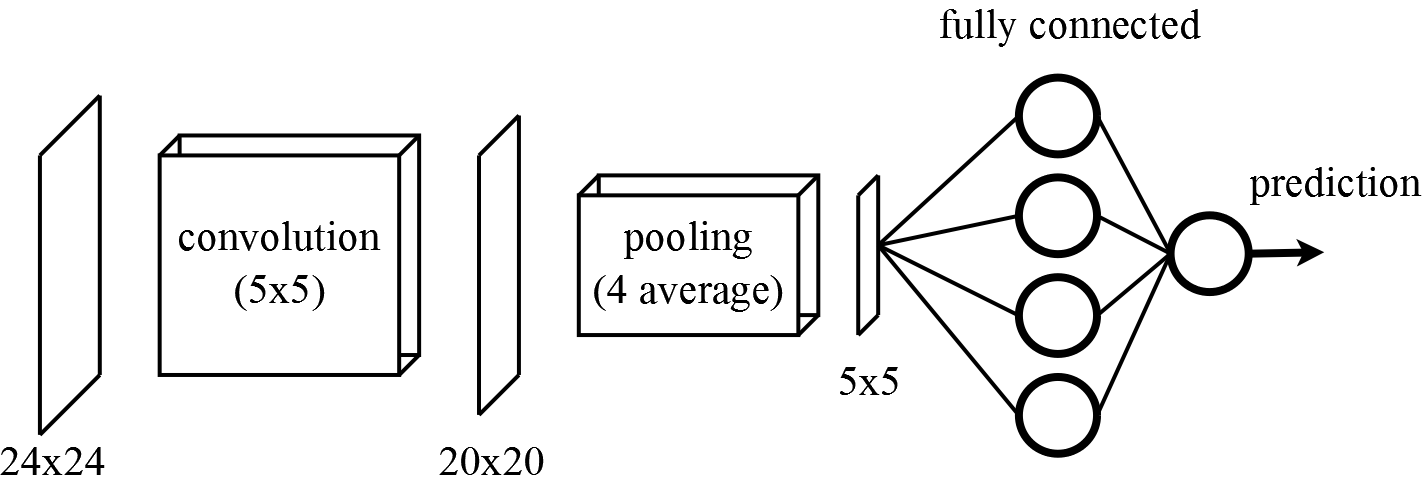}
	\end{center}
	\caption{
		The coarse position stage CNN. The first layer consists of the shared
		weights or convolution masks, which are summarized by the average
		pooling layer.
		Then a fully connected layer combines the features forwarded from the
		previous layer and delegates the final rating to a single perceptron.
	}
	\label{fig:struccoarse}
\end{figure}

The core architecture of the first stage CNN is summarized in
\figref{fig:struccoarse}.
The first layer is a convolutional layer with kernel size $5\times5$ pixels, one
pixel stride, and no padding.
The convolution layer is followed by an average pooling layer with window size
$4\times4$ pixels and four pixels stride, which is connected to a
fully-connected layer with depth one.
The output is then fed to a single perceptron, responsible for yielding the
final rating within the interval [0,1].
We have evaluated this architecture for different amounts of filters in the
convolutional layer and varying numbers of perceptrons in the fully connected
layer; these values are reported in \secref{sec:eval}.
The main idea behind the selected architecture is that the convolutional layer
learns basic features, such as edges, approximating the pupil structure.
The average pooling layer makes the CNN robust to small translations and
blurring of these features (e.g., due to the initial downscaling of the input
image).
The fully connected layer incorporates deeper knowledge on how to combine the
learned features for the coarse detection of the pupil position by using the
logistic activation function to produce the final
rating.

\subsection{Fine Positioning Stage}

Although the first stage yields an accurate pupil position estimate, it lacks
precision due to the inherent error introduced by the downscaling step.
Therefore, it is necessary to refine this estimate.
This refinement could be attempted by applying methods similar to those
described in \secref{sec:related} to a small window around the coarse pupil
position estimate.
However, since most of the previously mentioned challenges are not alleviated by
using this small window, we chose to use a second CNN that evaluates subregions
surrounding the coarse estimate in the original image.

The second stage CNN employs the same architecture pattern as the first stage
(i.e., convolution $\Rightarrow$ average pooling $\Rightarrow$ fully connected
$\Rightarrow$ single logistic perceptron) since their motivations are analogous.
Nevertheless, this CNN operates on a larger input resolution to provide
increased precision.
Intuitively, the input image for this CNN would be $96\times96$ pixels: the input
size of the first CNN input ($24\times24$) multiplied by the downscaling factor
($4$).
However, the resulting memory requirement for this size was larger than
available on our test device; as a result, we utilized the closest working size
possible: $89\times89$ pixels.
The size of the other layers were adapted accordingly.
The convolution kernels in the first layer were enlarged to $20\time20$ pixels to
compensate for increased noise and motion blur.
The dimension of the pooling window was increased by one pixel on each side,
leading to a decreased input size on the fully connected layer and reduced
runtime.
This CNN uses eight convolution filters and eight perceptrons due to the
increased size of the convolution filter and the input region size.
Subregions surrounding the coarse pupil position are extracted based on a window
of size $89\times89$ pixels centered around the coarse estimate, which is
shifted from $-10$ to $10$ pixels (with a one pixel stride) horizontally and vertically.
Analogously to the first stage, the center of the region with the highest CNN
rating is selected as fine pupil position estimate.

Despite higher computational costs in the second stage, our approach is highly
efficient and can be run on today's conventional mobile computers.

\subsection{CNN Training Methodology}
\label{subsec:training}

Both CNNs were trained using supervised batch gradient
descent~\cite{lecun2012efficient} (which is explained in detail in the
supplementary material) with a fixed learning rate of one.
Unless specified otherwise, training was conducted for ten epochs with a batch
size of 500.
Obviously, these decisions are aimed at an adaptive solution, where the training
time must be relatively short (hence the small number of epochs and high
learning rate) and no normalization (e.g., PCA whitening, mean extraction) can
be performed.
All CNNs' weights were initialized with random samples from a uniform
distribution, thus accounting for symmetry breaking.

While stochastic gradient descent searches for minima in the error plane more
effectively than batch learning~\cite{heskes1993line,orr1995dynamics} when give
valid examples, it is vulnerable to disastrous hops if given inadequate examples
(e.g., due to poor performance of the traditional algorithm).
On the contrary, batch training dilutes this error.
Nevertheless, we explored the impact of stochastic learning (i.e., using a batch
size of one) as well as an increased number of training epochs in
\secref{sec:eval}.

\subsubsection{Coarse Positioning CNN}

\begin{figure}[h]
	\begin{center}
		\includegraphics[width=1.0\linewidth]{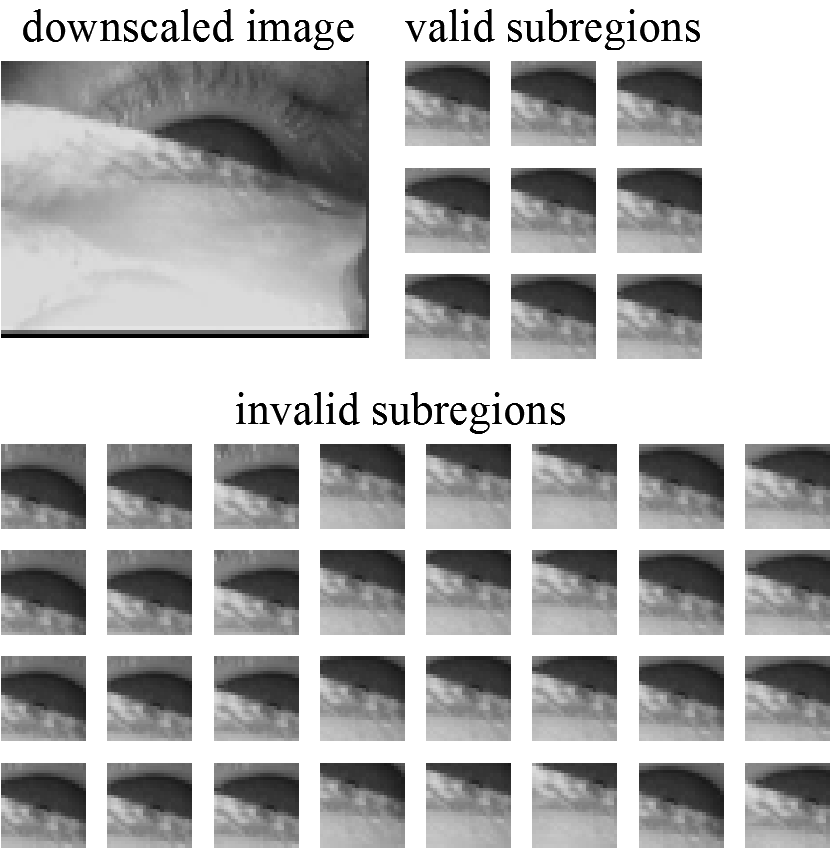}
	\end{center}
	\caption{
		Nine valid (top right) and 32 invalid (bottom) training samples for the coarse
		position CNN extracted from a downscaled input image (top left).
	}
	\label{fig:traindata}
\end{figure}

The coarse position CNN was trained on subregions extracted from the downscaled
input images that fall into two different data classes: containing a valid
($label=1$) or invalid ($label=0$) pupil center.
Training subregions were extracted by collecting all subregions with center
distant up to five pixels from the hand-labeled pupil center. Subregions with
center distant up to one pixel were labeled as valid examples while the
remaining subregions were labeled as invalid examples. As exemplified
by~\figref{fig:traindata}, this procedure results in nine valid and 32 invalid
samples per hand-labeled data.

We generated two training sets for the CNN responsible for the coarse position
detection.  The first training set consists of 50\% of the images provided by
related work from Fuhl et al.~\cite{fuhl2015excuse}.
The remaining 50\% of the images as well as the additional data sets from this
work~(see~\secref{sec:newdata}) are used to evaluate the detection performance
of our approach.
The purpose of this training is to investigate how the coarse positioning CNN
behaves on data it has never seen before.
The second training includes the first training set and 50\% of the images of
our new data sets and is employed to evaluate the complete proposed method (i.e.,
coarse and fine positioning).

\subsubsection{Fine Positioning CNN}

The fine positioning CNN (responsible for detecting the exact pupil position) is
trained similarly to the coarse positioning one.
However, we extract only one valid subregion sample centered at the hand-labeled
pupil center and eight equally spaced invalid subregions samples centered five
pixels away from the hand-labeled pupil center.
This reduced amount of samples per hand-labeled data relative to the coarse
positioning training is to constrain learning time, as well as main memory and
storage consumption.
We generated samples from 50\% of the images from
Fuhl~et~al.~\cite{fuhl2015excuse} and from the additional data sets.
Out of these samples, we randomly selected 50\% of the valid examples and 25\%
of the generated invalid examples for training.

\section{Data Set}
\label{sec:newdata}

\begin{figure}[h]
	\begin{center}
		\includegraphics[width=\columnwidth]{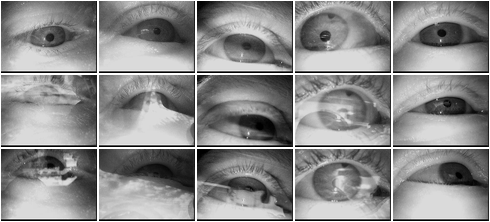}
		\caption{
			Samples from the additional data sets employed in this work.
			Each column belongs to a distinct data set.
			The top row includes non-challenging samples, which can be
			considered relatively similar to laboratory conditions and represent
			only a small fraction of each data set.
			The other two rows include challenging samples with artifacts caused
			by the natural environment.
		}
		\label{fig:newdatasets}
	\end{center}
\end{figure}

In this study, we used the extensive data sets introduced by
Fuhl~et~al.~\cite{fuhl2015excuse}, complemented by five additional hand-labeled
data sets.
Our additional data sets include 41,217 images collected during driving sessions
in public roads for an experiment \cite{kasneci2014driving} that was not related
to pupil detection and were chosen due the non-satisfactory performance of the
proprietary pupil detection algorithm.
These new data sets include fast changing and adverse illumination, spectacle
reflections, and disruptive physiological eye characteristics (e.g., dark spot
on the iris); samples from these data sets are shown in
\figref{fig:newdatasets}.


\section{Evaluation}
\label{sec:eval}
\label{sec:expeval}

Training and evaluation were performed on an
Intel\textsuperscript{\textregistered} Core\texttrademark i5-4670 desktop
computer with 8GB RAM. This setup was chosen because it provides a
performance similar to systems that are usually provided by eye-tracker vendors,
thus enabling the actual eye-tracking system to perform other experiments along with the evaluation.
The algorithm was implemented using MATLAB (r2013b) combined with the \emph{deep
learning toolbox}~\cite{palm2012prediction}.
During this evaluation, we employ the following name convention: K$_{n}$ and
P$_{k}$ signify that the CNN has \emph{n} filters in the convolution layer and
\emph{k} perceptrons in the fully connected layer.
We report our results in terms of the average pupil detection rate as a function of pixel distance between the algorithmically established and the hand-labeled pupil center.
Although the ground truth was labeled by experts in eye-tracking research,
imprecision cannot be excluded. Therefore, the results are discussed for a pixel error of five (i.e., pixel distance between the algorithmically established and the hand-labeled pupil center), analogously to~\cite{fuhl2015excuse,
swirski2012robust}.

\subsection{Coarse Positioning}

We start by evaluating the candidates from~\tblref{tbl:coarse} for the coarse
positioning CNN. All candidates follow the core architecture presented
in~\secref{subsec:coarsestage} and each candidate has a specific number of filters in the
convolution layer and perceptrons in the fully connected layer.
Their names are prefixed with \emph{C} for \emph{coarse} and, as
previously specified in~\secref{subsec:training}, were trained for ten epochs
with a batch size of 500.
Moreover, since \ckp{8}{16} provided the best trade-off between performance and
computational requirements, we chose this configuration to evaluate the impact
of training for an \emph{extended} period of time, resulting in the CNN \cext,
which was trained for a hundred epochs with a batch size of 1000.
Because of inferior performance and for the sake of space  we omit the results for the stochastic gradient descent learning but will make them available online.
\begin{table}
	\small
	\begin{center}
		\begin{tabular}{lcc}
			\toprule
			\textbf{CNN}
			& \begin{tabular}{@{}c@{}}
				\textbf{Conv.}\\
				\textbf{Kernels}
				\end{tabular}
			& \begin{tabular}{@{}c@{}}
				\textbf{Fully Conn.}\\
				\textbf{Perceptrons}
				\end{tabular}\\
			\midrule
			\ckp{4}{8}  & \phantom{0}4 & \phantom{0}8 \\
			\ckp{8}{8}  & \phantom{0}8 & \phantom{0}8 \\
			\ckp{8}{16}  & \phantom{0}8 & 16 \\
			\ckp{16}{32} & 16           & 32 \\
			\bottomrule
		\end{tabular}
	\end{center}
	\caption{
		Evaluated configurations for the coarse positioning CNN (as described
		in~\secref{subsec:coarsestage}).
	}
	\label{tbl:coarse}
\end{table}

\begin{figure}[h]
	\begin{center}
		\includegraphics[trim={0.27cm 0 0.4cm 0},clip,width=\columnwidth]{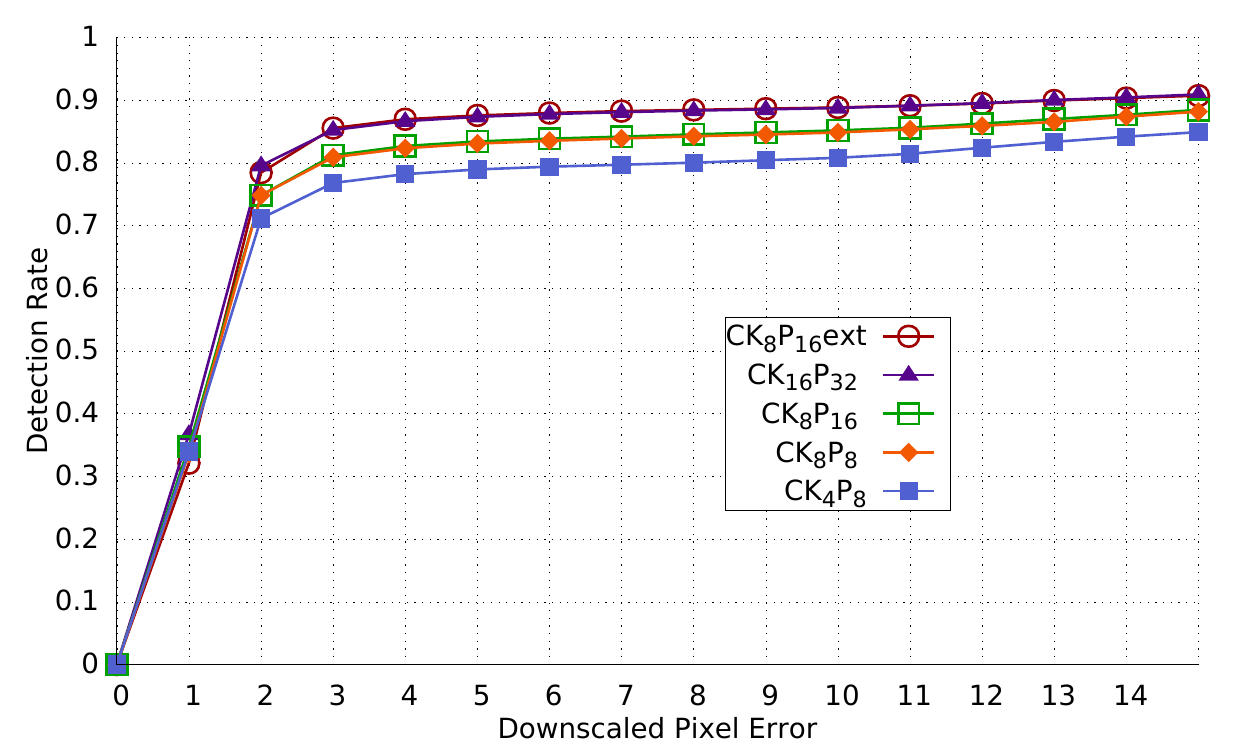}
	\end{center}
	\caption{
		Performance for the evaluated coarse CNNs trained on 50\% of images from
		all data sets and evaluated on all images from all data sets.
	}
	\label{fig:evalcoarse}
\end{figure}

\figref{fig:evalcoarse} shows the performance of the coarse positioning CNNs
when trained using 50\% of the images randomly chosen from all data sets and evaluated on all images.
As can be seen in this figure, the number of filters in the
first layer (compare \ckp{4}{8}, \ckp{8}{8}, and \ckp{16}{32}) and
extensive learning (see \ckp{8}{16} and \cext) have a higher impact than the number of perceptrons in the fully connected layer (compare
\ckp{8}{8}  to \ckp{8}{16}).
Moreover, these results indicate that the amount of filters in the convolutional
layer still has not been saturated (i.e., there are still high level features
that can be learned to improve accuracy). However, it is important to notice
that this is the most expensive parameter in the proposed CNN architecture in
terms of computation time and, thus, further increments must be carefully included.

\begin{figure}[h]
	\begin{center}
		\includegraphics[trim={0.27cm 0 0.4cm 0},clip,width=\columnwidth]{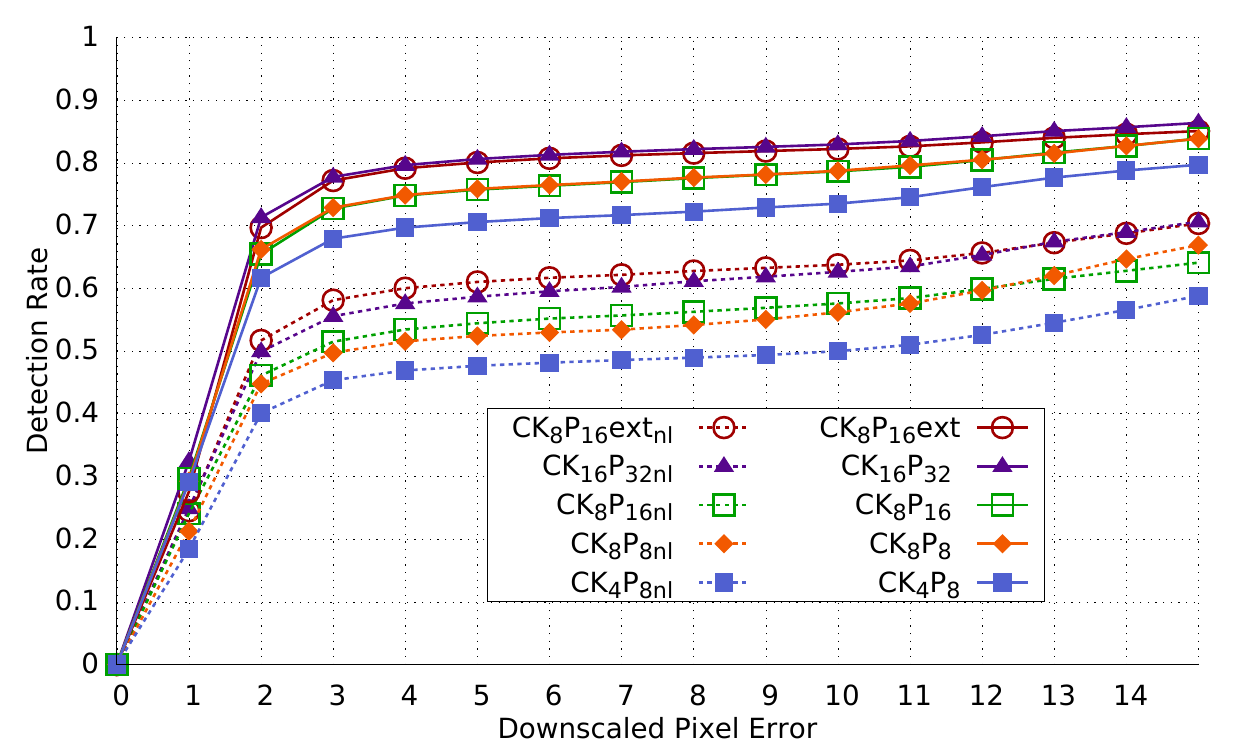}
	\end{center}
	\caption{
		Performance of coarse positioning CNNs on all images provided by this work.
		The solid lines belong to the CNNs trained on 50\% of all images from
		all data sets.
		The dotted lines belong to the \emph{nl}-CNNs, which were trained only
		on 50\% of the data set provided by Fuhl et al.~\cite{fuhl2015excuse}
		and, thus, have \emph{not learned} on examples from the evaluation data
		sets.
	}
	\label{fig:evalcoarsenl}
\end{figure}

To evaluate the performance of the coarse CNNs only on data they have not seen before,
we additionally retrained these CNNs from scratch using only 50\% of the images
in the data sets provided by Fuhl et al.~\cite{fuhl2015excuse} and evaluated
their performance solely on the new data sets provided by this work.
These results were compared to those from the CNNs that were trained on 50\% of
images from all data sets and are shown in~\figref{fig:evalcoarsenl}.
The CNNs that have \emph{not learned} on the new data sets are identified by the
suffix \emph{nl}.
All \emph{nl}-CNNs exhibited a similar decrease in performance relative to their
counterparts that have been trained on samples from all the data sets. We hypothesize that this
effect is due to the new data sets holding new information (i.e., containing new
challenging patterns not present in the training data); nevertheless, the CNNs
generalize well enough to handle even these unseen patterns decently.

\subsection{Fine Positioning}

The \emph{fine} positioning CNN (\cfin) uses the
\ckp{8}{8} in the first stage and was trained on 50\% of the data from all data
sets.
Evaluation was performed against four state-of-the-art algorithms, namely,
\emph{ExCuSe}~\cite{fuhl2015excuse}, \emph{SET}~\cite{javadi2015set},
Starburst~\cite{li2005starburst}, and \'{S}wirski~\cite{swirski2012robust}.
Furthermore, we have developed two additional fine positioning methods for this
evaluation:
\begin{itemize}
	\item \cray: this method employs \ckp{8}{8} to determine a coarse
		estimation, which is then refined by sending \emph{rays} in eight equidistant
		directions with a maximum range of thirty pixels. The difference between
		every two adjacent pixels in the ray's trajectory is calculated, and the
		center of the opposite ray is calculated. Then, the mean of the two
		closest centers is used as fine position estimate. This method is used
		as reference for hybrid methods combining CNNs and traditional pupil
		detection methods.
	\item \csin: this method uses only a \emph{single} CNN similar to the ones
		used in the coarse positioning stage, trained in an analogous fashion.
		However, this CNN uses an input size of $25\times25$ pixels to obtain an
		even center. This method is used as reference for a parsimonious (although
		costlier than the original coarse positioning CNNs) single stage CNN
		approach and was designed to be employed on systems that cannot handle
		the second stage CNN.
\end{itemize}
For reference, the coarse positioning CNN used in the first stage of \cfin{} and
\cray{} (i.e., \ckp{8}{8}) is also shown.

\begin{figure*}[htbp]
	\begin{center}
		\subfloat[][\label{a}]{
			\includegraphics[trim={2cm 0 1.8cm 0},clip,width=.33\linewidth]{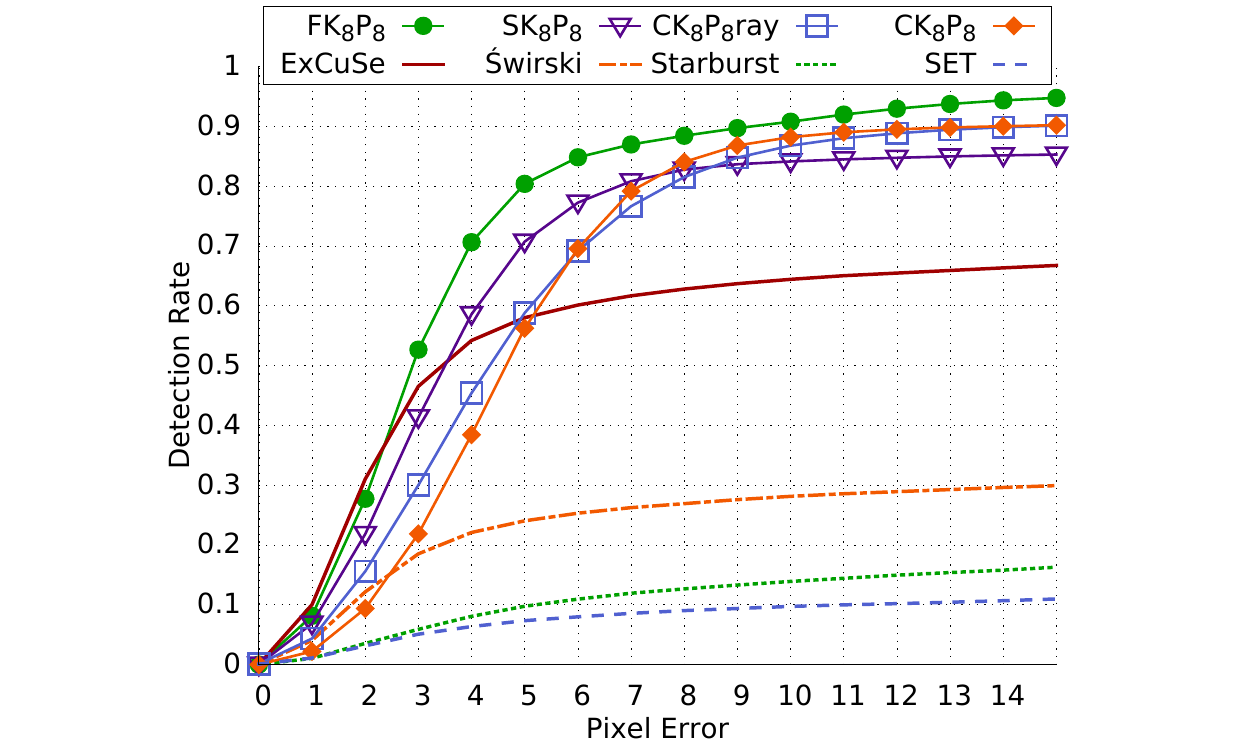}
			\label{fig:evalfineex}
		}
		\subfloat[][\label{b}]{
			\includegraphics[trim={2cm 0 1.8cm 0},clip,width=.33\linewidth]{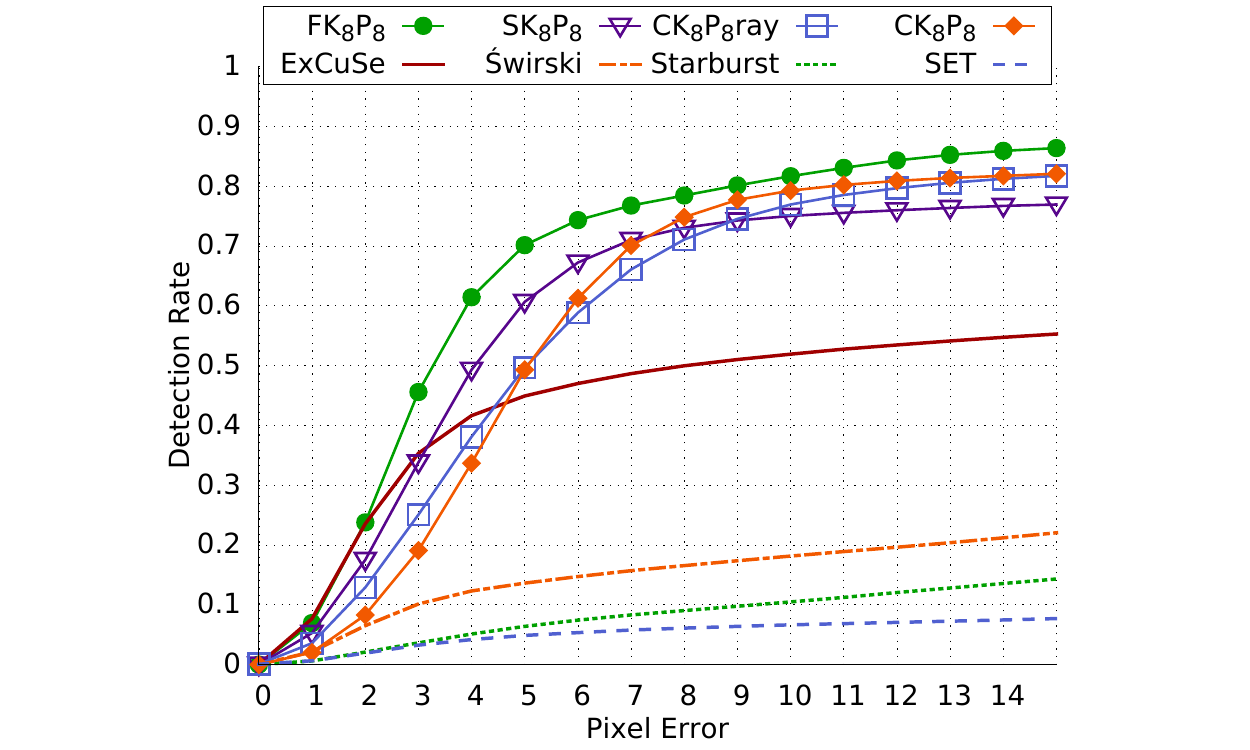}
			\label{fig:evalfine}
		}
		\subfloat[][\label{c}]{
			\includegraphics[trim={2cm 0 1.8cm 0},clip,width=.33\linewidth]{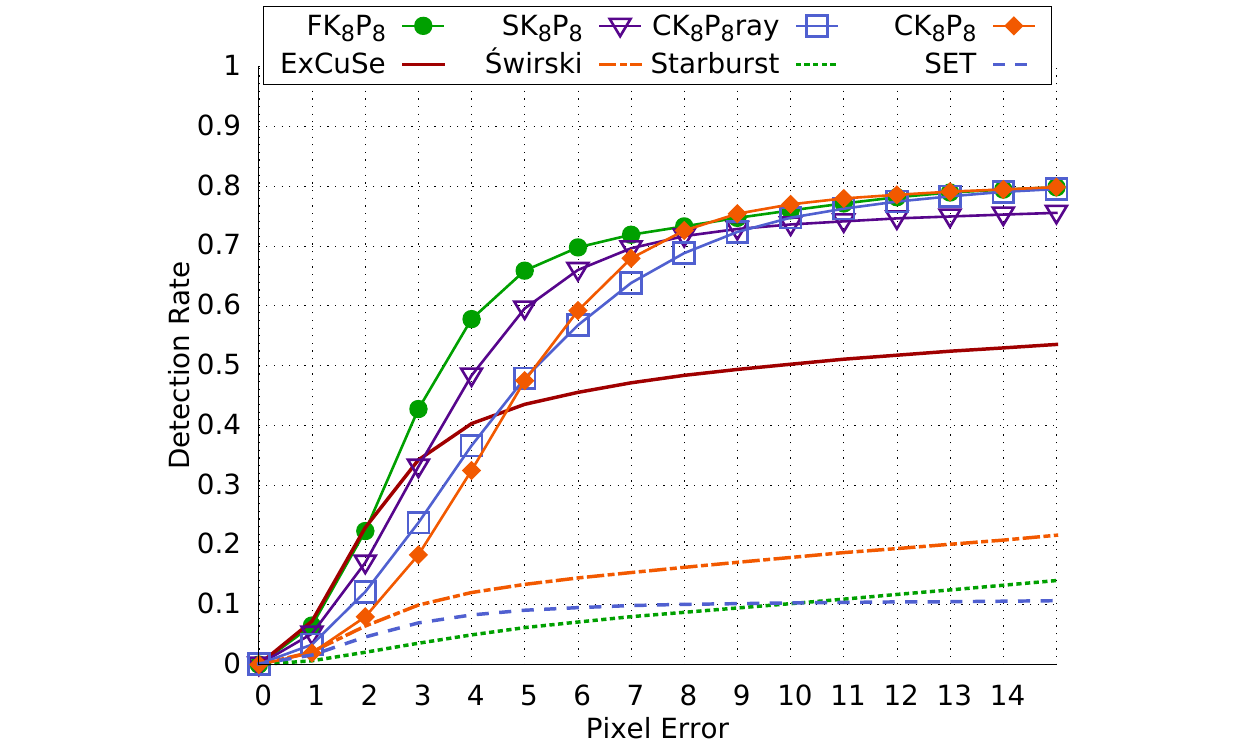}
			\label{fig:evalfineunseen}
		}
	\end{center}
	\caption{
		All CNNs were trained on 50\% of images from all data sets.
		Performance for the selected approaches on (a) all images from the data
		sets from~\cite{fuhl2015excuse}, (b) all images from all data sets, and
		(c) all images not used for training from all data sets.
	}
\end{figure*}

All CNNs in this evaluation were trained on 50\% of the images
randomly selected from all data sets.
To avoid biasing the evaluation towards the data set introduced by this work, we
considered two different evaluation scenarios.
First, we evaluate the selected approaches only on images from the data sets
introduced by Fuhl et al.~\cite{fuhl2015excuse}, and, in a second step, we
perform the evaluation on all images from all data sets.  The outcomes are shown
in the Figures~\ref{fig:evalfineex}~and~\ref{fig:evalfine}, respectively.
Finally, we evaluated the performance on all images not used for training from
all data sets.
This provides a realistic evaluation strategy for the aforementioned adaptive solution. The results are shown in \figref{fig:evalfineunseen}.

In all cases, both \cfin{} and \csin{} surpass the best performing state-of-the-art
algorithm by approximately 25\% and 15\%, respectively.
Moreover, even with the penalty due to the upscaling error, the evaluated coarse
positioning approaches mostly exhibit an improvement relative to the
state-of-the-art algorithms.
The hybrid method \cray, did not display a significant improvement relative to
the coarse positioning; as previously discussed, this behavior is expected as
the traditional pupil detection methods are still afflicted by aforementioned
challenges, regardless of the availability of the coarse pupil position
estimate.
Although the proposed method (\cfin) exhibits the best pupil detection rate, it
is worth highlighting the performance of the \csin{} method combined with
its reduced computational costs.
Without accounting for the downscaling operation, \csin{} has an operating cost
of eight convolutions ($(6\times6) * (20\times20) * 8 = 115200$ FLOPS) plus
eight average-pooling operations ($(20\times20) * 8 = 3200$ FLOPS) plus
$(5\times5\times8)*8 = 1600$ FLOPS from the fully connected layer and $8$
FLOPS from the last perceptron, totaling $120008$ FLOPS per run.
Given an input image of size $96\times72$ and the input size of $25\times25$,
$72\times48=3456$ runs are necessary, requiring $\approx415\times10^6$ FLOPS
without accounting for extra operations (e.g. load/store).
These can be performed in real-time even on the accompanying eye tracker system
CPU, which yields a baseline 48 GFLOPS~\cite{intel3540}.

\subsection{CNN Operational Behavior Analysis}

To analyze the patterns learned by the CNNs in detail, we further inspected the
filters and weights of \ckp{8}{8}. Notably, the CNNs had learned similar filters and
weights.
Representatively, we chose to report based on \ckp{8}{8} since it can be more easily
visualized due to its reduced size.

\begin{figure}[h]
	\begin{center}
		\includegraphics[width=\columnwidth]{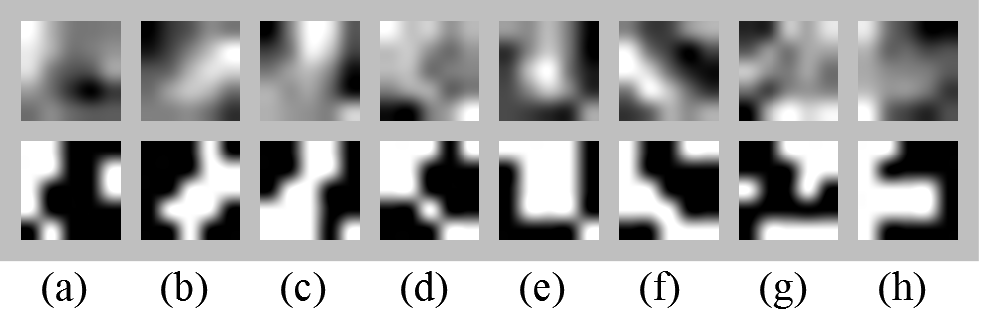}
	\end{center}
	\caption{
		\ckp{8}{8} filters in the convolutional layer. The first row displays
		the intensity of the filters' weights, and the second row indicates
		whether the weight was positive (white) or negative (black).
		For visualization, the filters were resized by a factor of twenty using
		bicubic interpolation and normalized to the range [0,1].
	}
	\label{fig:learnedfilters}
\end{figure}

\begin{figure}[h]
	\begin{center}
		\includegraphics[width=1.0\linewidth]{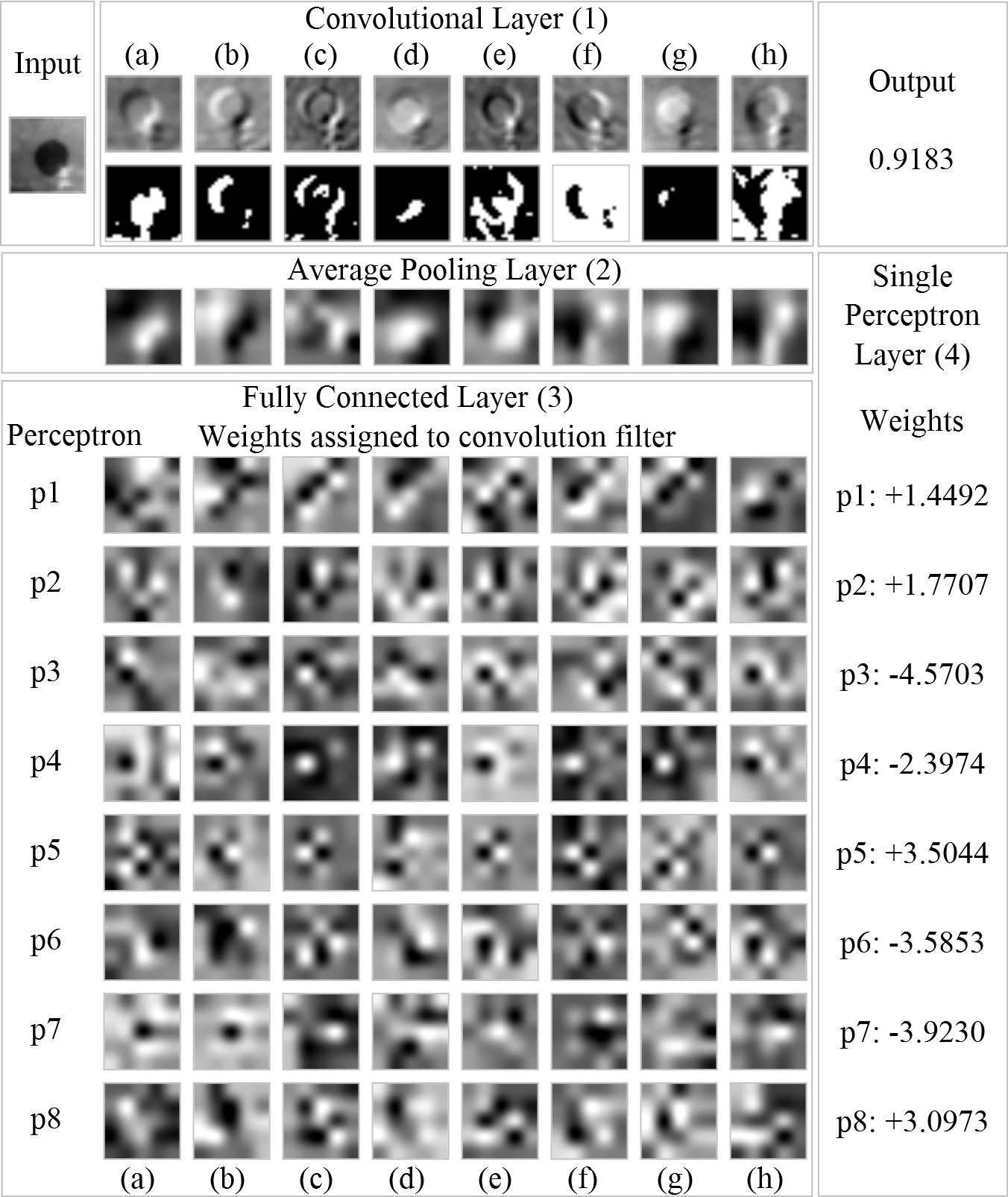}
	\end{center}
	\caption{
		\ckp{8}{8} response to a valid subregion sample.
		The convolutional layer includes the filter responses in the first row
		and positive (in white) and negative (in black) responses in the second
		row.
		The fully connected layer shows the weight maps for each
		perceptron/filter pair.
		For visualization, the filters were resized by a factor of twenty using
		bicubic interpolation and normalized to the range [0,1].
		The filter order (i.e., (a) to (b) ) matches that of
		\figref{fig:learnedfilters}.
	}
	\label{fig:learnedweight}
\end{figure}

The first row of \figref{fig:learnedfilters} shows the filters learned in the
convolution layer, whereas the second row shows the sign of these filters'
weights where white and black represent positive and negative weights,
respectively.
Filter (e) resembles a center surround difference, and the remaining filters
contain round shapes, most probably performing edge detections.
It is worth noticing that the filters appear to have developed in complementing
pairs (i.e., a filter and its inverse) to some extent.
This behavior can be seen in the pairs (a,c), (b,d), and (f,g), while filter
(e) could be paired with (h) if the latter further develops its top and bottom
right corners.
Furthermore, the convolutional layer response based on these filters when given
a valid subregion as input is demonstrated in \figref{fig:learnedweight}. The first row displays the filters responses, and the second row shows positive
(white) and negative (black) responses.

The weights of all perceptrons in the fully connected layer are also displayed in
\figref{fig:learnedweight}.
In the fully connected layer area, the first column identifies the perceptron in
the fully connected layer (i.e., from p1 to p8), and the other columns display
their respective weights for each of the filters from
\figref{fig:learnedfilters} (i.e., from (a) to (h) ).
Since the output weight assigned to perceptrons p1, p2, p5, and p8 are
positive, these patterns will respond to centered pupils, while the opposite
is true for perceptrons p3, p4, p6, and p7 (see the single perceptron layer in
\figref{fig:learnedweight}).
This behavior is caused by the use of a \emph{sigmoid} function with outputs
ranging between zero and one. If a \emph{tanh} function was used instead, it
could lead to negative weights being positive responses.
Based on the negative and positive values (\figref{fig:learnedfilters}, second
rows of the convolutional layer), the filters (b) and (f) display opposite
responses.
Based on the input coming from the average pooling layer, p2 displays the best
fitting positive weight map for the filter response (a) in the first row of the
convolutional layer. Similarly, p1 provides a best fit for (b) and (d), and both
p1 and p8 provide good fits for (c) and (f);
no perceptron presented an adequate fit for (g) and (h), indicating that these
filters  are possibly employed to respond to off center pupils.
Moreover, all negatively weighted perceptrons present high values at the center for the
response filter (e) (the possible center surrounding filter), which could  be
employed to ignore small blobs. In contrast, p5 (a positively weighted
perceptron) weights center responses high.

\section{Conclusion}
We presented a naturally motivated pipeline of specifically configured CNNs for
robust pupil detection and showed that it outperforms state-of-the-art
approaches by a large margin while avoiding high computational costs. For the
evaluation we used over 79.000 hand labeled images -- 41.000 of which were
complementary to existing images from the literature -- from real-world recordings with artifacts such as
reflections, changing illumination conditions, occlusion, etc. Especially for
this challenging data set, the CNNs reported considerably higher detection rates
than state-of-the-art techniques. Looking forward, we are planning to
investigate the applicability of the proposed pipeline to online scenarios,
where continuous adaptation of the parameters is a further challenge.


\end{document}